\documentclass[letterpaper]{article} 

\usepackage{aaai24}  

\usepackage{times}  
\usepackage{helvet}  
\usepackage{courier}  
\usepackage[hyphens]{url}  
\usepackage{graphicx} 
\usepackage{natbib}  
\usepackage{caption} 
\usepackage{algorithm}
\usepackage{algpseudocode}

\usepackage{newfloat}
\usepackage{listings}
\DeclareCaptionStyle{ruled}{labelfont=normalfont,labelsep=colon,strut=off} 
\floatstyle{ruled}
\newfloat{listing}{tb}{lst}{}
\floatname{listing}{Listing}
\title{Decision-BADGE: Decision-based Adversarial Batch Attack\\with Directional Gradient Estimation}
\author {
    Geunhyeok Yu    \textsuperscript{\rm 1},
    Minwoo Jeon     \textsuperscript{\rm 1},
    Hyoseok Hwang   \textsuperscript{\rm 1}
}
\affiliations {
    \textsuperscript{\rm 1}Department of Software Convergence, Kyung Hee University, Republic of Korea\\
    \{geunhyeok, alsdn2530, hyoseok\}@khu.ac.kr
}

\usepackage{bibentry}

\usepackage{lipsum}
\usepackage{import}
\usepackage{booktabs}
\usepackage{subcaption}
\usepackage[dvipsnames]{xcolor}

\usepackage{amsthm}
\usepackage{amsmath}
\usepackage{amssymb}

\usepackage{hyperref}
\usepackage{fontawesome}

\newtheorem{theorem}{Theorem}[section]
\newtheorem{lemma}[theorem]{Lemma}

\newcommand{\syminput}{\mathbf{x}}
\newcommand{\symgt}{\mathbf{y}}
\newcommand{\sympert}{\mathbf{p}}
\newcommand{\symniters}{T}
\newcommand{\symstep}{\mathbf{u}}
\newcommand{\symclassifier}{\mathcal{C}}
\newcommand{\symgrad}{\mathbf{g}}
\newcommand{\symgamma}{\gamma}

\begin{document}

\maketitle


\begin{abstract}
\label{sec:abstract}
The susceptibility of deep neural networks (DNNs) to adversarial examples has prompted an increase in the deployment of adversarial attacks.
Image-agnostic universal adversarial perturbations (UAPs) are much more threatening, but many limitations exist to implementing UAPs in real-world scenarios where only binary decisions are returned.
In this research, we propose Decision-BADGE, a novel method to craft universal adversarial perturbations for executing decision-based black-box attacks.
To optimize perturbation with decisions, we addressed two challenges, namely, the gradient's magnitude and direction. 
First, we use batch accuracy loss that measures the distance from distributions of ground truth and accumulating decisions in batches to determine the magnitude of the gradient. 
This magnitude is applied in the direction of the revised simultaneous perturbation stochastic approximation (SPSA) to update the perturbation. 
This simple, yet efficient method can be easily extended to score-based and targeted attacks. 
Experimental validation across multiple victim models demonstrates that the Decision-BADGE outperforms existing attack methods, even image-specific and score-based attacks. 
In particular, our proposed method shows a superior attack success rate with less training time. 
The research also shows that Decision-BADGE can successfully deceive unseen victim models and accurately target specific classes.
The code is available at our \textnormal{\href{https://github.com/AIRLABkhu/Decision-BADGE}{\faGithub{~GitHub}}}~\footnote{\texttt{https://github.com/AIRLABkhu/Decision-BADGE}} repository.
\end{abstract}

\section{Introduction}
\label{sec:introduction}

Deep Neural Networks (DNNs) are considered among the most versatile and sophisticated machine learning architectures. 
Optimization algorithms refine the networks' parameters by autonomously identifying the optimal decision boundaries.
For this reason, revolutionary progress has been achieved in numerous computer vision tasks~\cite{guo2018review, liu2017survey}.
However, it has been proven that DNNs are highly vulnerable to adversarial examples~(AEs), which are indistinguishable from the original image by adding a tiny amount of adversarial perturbation~\cite{Szegedy_2014_ICLR}. 
This can be critical, as adversarial attacks using AEs threaten the safety of DNN-based applications. They can confuse the networks~\cite{akhtar2018threat}, compromise privacy~\cite{Zhang_2020_CVPR,Wang_2021_NIPS}, duplicate or steal a model~\cite{Chandrasekaran_2020_USENIX_Security,Jagielski_2020_USENIX_Security}, and intentionally manipulate the model's decisions~\cite{williams2023black}.

\begin{figure}[t!]
    \begin{center}
        \begin{subfigure}{0.93\linewidth}
            \includegraphics[width=\linewidth]{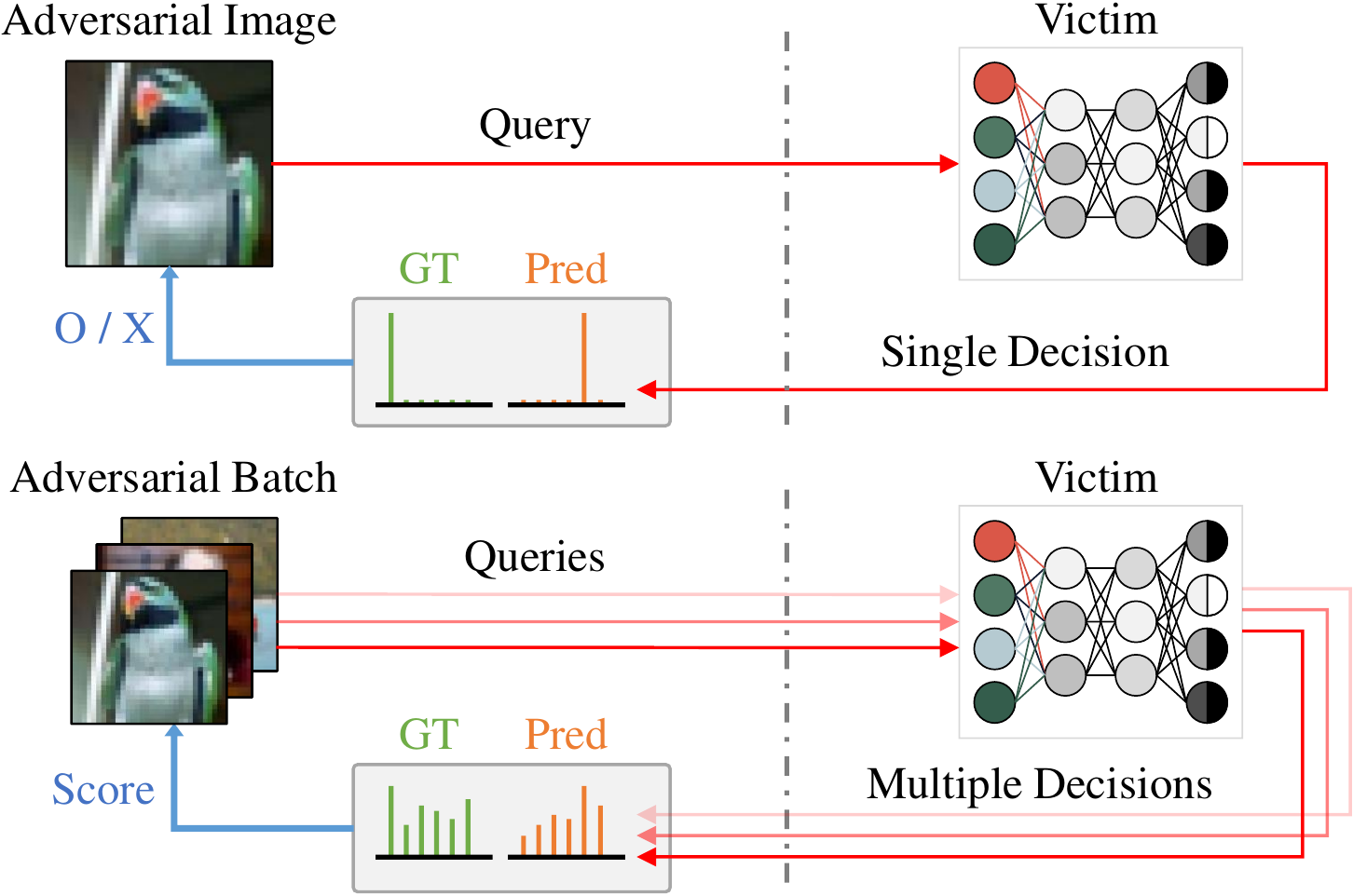}
            \caption{Decision-based Batch Accuracy Loss}
            \label{subfig:batch_attack-uap}
        \end{subfigure}
        \vspace{5mm}

        \begin{subfigure}{0.93\linewidth}
            \includegraphics[width=\linewidth]{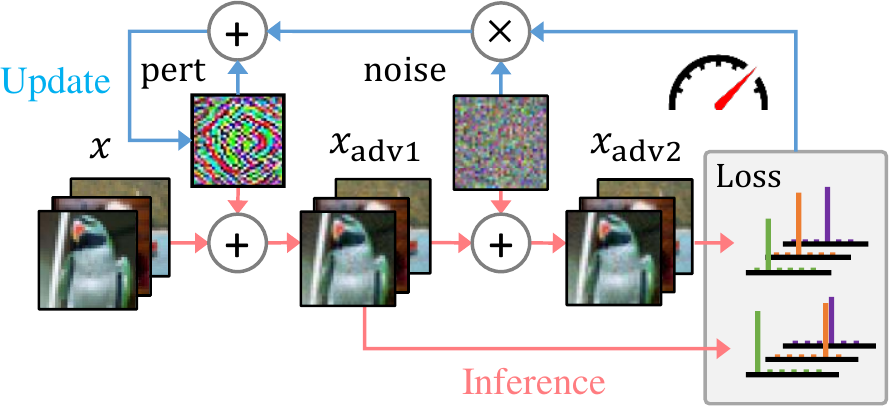}
            \caption{An Overview of Decision-BADGE}
            \label{subfig:batch_attack-ours}
        \end{subfigure}
    \label{fig:batch_attack}
    \caption{(a) shows the difference between the binary decision loss and the proposed batch accuracy loss that aggregates multiple binary decisions to form a continuous score. (b) illustrates the Decision-BADGE framework that utilizes the SPSA algorithm with the batch accuracy loss.}
    \end{center}
\end{figure}

Since the introduction of Universal Adversarial Perturbation (UAP) by Moosavi-Dezfooli et al.~\cite{Moosavi-Dezfooli_2017_CVPR}, adversarial attacks have become even more menacing. 
AEs can be generated from any input image by simply applying an image-agnostic perturbation, allowing the UAP to capture the entire decision boundary of a victim model~\cite{Zirui_2022_CVPR}. 
Unlike image-dependent methods, when generating an AE for an input image, once the UAP is created, it only needs to be added to the input image, ensuring that AEs are generated promptly.
Furthermore, this fooling technique is easily transferable across other network models.

Fortunately, adversarial attacks have yet to cause critical problems in real-world scenarios. 
The main reason is that in the real world, the model is unknown, so only black-box attacks are possible. 
Therefore, the white-box approaches with the superior performance of optimizing perturbations via backpropagation through the model are no longer available~\cite{Cheng_2018_arXiv,Cheng_2019_NIPS,Huang_2019_arXiv}.  
In addition, score-based black-box attacks, which use the confidence scores or the probability distribution over the classes for a query, cannot be applied to services that provide binary decisions or the top-1 class label only. 

Recently, methods for finding decision boundaries in a black box with a few queries have been studied~\cite{brendel2017decision, chen2020hopskipjumpattack}.
While these methods may be suitable for real-world scenarios, they are not appropriate for crafting UAPs, as such techniques generate image-specific perturbations.
Another approach to craft UAPs in a decision-based black box attack is through transfer-based attacks~\cite{cheng2019improving}, which utilizes the transferability of AEs across multiple models.
However, the success rate of transferred AEs tends to be lower, particularly when there is a significant architectural gap between the models, such as between convolution-based and transformer-based models.

Zeroth-order optimization methods also play an essential role in fooling the model using only decisions.
Chen et al.~\cite{Jinyin_2019_CnS} and Wu et al.~\cite{Chenwang_2021_CEC} employed a genetic algorithm to search for an adversarial perturbation. 
Random Gradient-Free~(RGF)~\cite{Nesterov_2017_Foundations-of-Computational-Mathematics}, and Simultaneous Perturbation Stochastic Approximation~(SPSA)~\cite{Spall_TAC_1992} are well designed gradient-free optimization algorithms that can be applied to black-box adversarial attack~\cite{Cheng_2019_NIPS,Cheng_2018_arXiv,Huang_2019_arXiv}.
However, utilizing zeroth-order optimization with a single decision per update may introduce noise into the optimization steps, potentially slowing down the entire optimization process.

In this paper, we propose \textit{\textbf{Decision}-based \textbf{B}atch \textbf{A}ttack with \textbf{D}irectional \textbf{G}radient \textbf{E}stimation (Decision-BADGE)}, which aims to efficiently craft universal adversarial perturbations in the decision-based black-box attacks. 
We propose the batch accuracy loss, based on the Hamming distance, to precisely measure the loss magnitude from binary decisions, which accumulates the distance between distributions on a mini-batch as illustrated in Figure ~\ref{subfig:batch_attack-uap}.
The accuracy loss is applied to the revised SPSA and utilized to determine the magnitude of the update~(see Figure ~\ref{subfig:batch_attack-ours}).
The proposed method is effectively extensible, allowing other distance metrics and attack methods, such as targeted attacks or score-based attacks, to be easily applied.
The proposed method  achieves a white-box level attack success rate with a similar number of perturbation updates.
The main contributions of this study can be summarized as follows:

\begin{itemize}
    \setlength\itemsep{0.5mm}
    \item 
    In a decision-based black-box attack, we only know the decision for a query. 
    Thus, we propose a novel method to improve the optimization performance using the batch accuracy loss with a distribution of decisions. 
    \item We applied a new optimization technique that takes advantage of the SPSA algorithm and Adam optimization and demonstrated the superiority over other combinations of optimization algorithms. 
    \item We mathematically formulated the loss function for the accuracy loss based on the Hamming distance and compared it with other existing loss functions.
    \item Our method overperforms other methods in terms of training time efficiency while achieving white-box level ASR.
\end{itemize}

\section{Related Works}
\label{sec:related_works}


\subsection{Universal Adversarial Attack}
\label{ssec:related_works-universal_attack}

Traditional adversarial attack methods are image-dependent attacks. 
It refers to exploiting the victim model by optimizing perturbation for one image. 
UAP based on DeepFool (DF-UAP)~\cite{moosavi2016deepfool} was introduced in which a perturbation can be applied to any input image in contrast to image-dependent perturbations by iterative boundary search~\cite{Moosavi-Dezfooli_2017_CVPR}. 
Singular value-based UAP~\cite{Khrulkov_CVPR_2018}, which is relatively data-efficient to build compared to UAP, was demonstrated.
Network Adversary Generation~(NAG)~\cite{Mopuri_2018_CVPR} is an adversarial example generator based on Generative Adversarial Networks (GAN)~\cite{Goodfellow_2014_NIPS}. 
They demonstrated that generators can capture the perturbation geometry and achieve high fooling transferability. 
Generative Adversarial Perturbation (GAP) is another GAN-based adversarial perturbation generator~\cite{Poursaeed_2018_CVPR}.
NAG and GAP are capable of generating not only image-dependent adversaries but also UAPs. 

\subsection{Black-box Adversarial Attack}
\label{ssec:related_works-bb_attack}

The attacker cannot access the victim's architecture, gradient, or training process in a black-box attack.
We classified some approaches by the provided information to the attacker. 

\noindent
\textbf{Transfer-based methods.} 
Transfer-based methods do not directly access the victims but require a substitute model. 
Local Substitute Network~\cite{Papernot_2017_CCS} trained perturbations using a known network and attacked an unknown victim model. 
Curls and Whey~\cite{Yucheng_2019_CVPR} is another black-box attack method using a substitute network. 
The authors tried to boost the attack process by finding a faster trajectory to make the data point to cross the decision boundary. 
Translation-Invariant Attack~\cite{Yinpeng_2019_CVPR} tackled true translation-invariance to attack convolutional neural networks properly. 

\noindent
\textbf{Score-based methods.} 
These methods take the rich confidence score for each query. 
Zeroth-order Optimization (ZOO)~\cite{Chen_2017_CCS} proposed a black-box attack method by optimizing randomly selected pixels using Newton's Method.
Simple Black-box Adversarial Attacks~(SimBA)~\cite{Chuan_2019_ICML} is another powerful black-box attack method. 
They proposed to use a set of orthonormal vectors as a direction set for local search. 

\noindent
\textbf{Decision-based methods.} 
In a real-world scenario, the victim service provides decisions only.
Therefore, decision-based attacks have been researched from the perspective of query efficiency. 
RGF optimization was applied to solve this problem~\cite{Ghadimi_2013_SIAM,Cheng_2018_arXiv}.
The local search algorithm was also applied~\cite{Brunner_2019_CVPR} but they further applied the Biased Boundary Attack to build low-frequency perturbations. 
Reliable attack precisely selects the magnitude of an update~\cite{Brendel_2018_ICLR,Chen_2020_SSP}.
The genetic algorithm was proposed to be applied, finding the optimal step on a sparse decision space~\cite{Vo_2022_ICLR}.
Wu et al. introduced Decision-based Universal Attack~(DUAttack), which is an algorithm to build a universal perturbation using decisions~\cite{Wu_2020_arXiv}.
They aggregated multiple images into a mini-batch to properly update the perturbations.
To the best of our knowledge, this was the only successful approach to solving the decision-based universal attack problem.

\subsection{Zeroth-order Optimization}
\label{ssec:related_works-zoo}

Several methods have been studied to apply gradient-based optimization to craft adversarial perturbations without the gradient of victim models. 
Natural Evolutionary Strategy~(NES) was employed to adversarial attack to optimize without calculating gradients~\cite{Salimans_arXiv_2017,Ilyas_2018_PMLR}.
Some researchers addressed this problem by employing the RGF algorithm~\cite{Nesterov_2017_Foundations-of-Computational-Mathematics}.
Most adversarial attacks based on the RGF aim to identify effective random steps. 
Randomly sampled steps on a hypersphere can be used as an update with the momentum optimization method~\cite{Cheng_2019_NIPS}.
The generator was also employed in zeroth-order optimization to generate a step using neural networks~\cite{Huang_2019_arXiv}.
These methods effectively addressed score-based attack problems using the RGF. 
SPSA is another method for solving zeroth-order optimization problems~\cite{Spall_TAC_1992,Maryak_ACC_2001}. 
It can be applied to address black-box adversarial attack problems~\cite{Ilyas_arXiv_2017,Uesato_2018_PMLR}. 
SPSA with gradient correction (SPSA-GC)~\cite{Oh_2023_CVPR} improved the direction of the steps in the black-box prompt tuning task by integrating the SPSA algorithm with the Nesterov Accelerated Gradient optimization algorithm~\cite{Nesterov_1983_AMF}.
Thus far, the methods NES, RGF, and SPSA have proven unsuitable for decision-based attacks, as they are unable to accurately evaluate the direction, particularly in the context of universal attacks.
\section{Methods}
\label{sec:methods}
In this section, we introduce Decision-BADGE, a decision-based method incorporating a revised version of the SPSA algorithm. 
This revised version aggregates decisions in a batch to recover the lost distribution. We propose the accuracy loss function, a simple, yet highly effective tool for comparing two distributions. 
Furthermore, we provide mathematical proof to demonstrate the feasibility of our loss function.

Model fooling is to confuse a victim classifier to make a misclassification by adding a tiny amount of perturbation to the input. 
Let the original input image, the perturbation vector, and the classifier be $\syminput_i, \sympert_i \in \mathbb{R}^{N_\text{in}}$ and $\symclassifier: \mathbb{R}^{N_\text{in}} \mapsto [0,1]^{N_\text{cls}}$, respectively. 
As long as our objective is to make UAPs, one perturbation should be able to fool as many images as possible. 
Moreover, in the real-world scenario, the attacker can only acquire the victim's decision. 
Therefore, our objective is formulated following:
\begin{equation}
    \label{eq_fooling_agnostic}
    \operatorname{min}{\sum_{i=1}^{N}\langle \mathbb{D}(\syminput_i), \mathbb{D}(\syminput_i + \sympert) \rangle}~\text{subject to:} \quad \|\sympert\|_{\infty}\le \epsilon
\end{equation}
for $N$ input images, where $\langle\cdot,\cdot\rangle$ is inner product between two vectors and $\|\cdot\|_{l}$ is $L_{l}$-norm of a vector, $\mathbb{D}(\syminput)$ denotes the one-hot vector corresponding to the top-1 element of $\symclassifier(\syminput)$.

\subsection{Batch Accuracy Loss}
\label{ssec:methods-adversarial_batchattack}

The decision of the victim model forgets score distribution.
This hinders the calculation of losses by the difference between the distributions. 
Our crucial concept involves computing and updating the distribution in mini-batches rather than deriving it from a single image.
This allows us to calculate the distance of two batches' decision distributions. 
There are several distance metrics for distribution, such as Cross Entropy (CE), Kullback-Leibler Divergence (KLD), Earth Mover's Distance (EMD), and the Hamming distance. 
However, not all metrics are suitable for addressing this particular issue.
When attempting to apply SPSA-based optimization, the loss function must be convex to prevent getting trapped in local minima, and it should be Lipschitz-continuous to ensure stable optimization~\cite{Yu_2021_TAC}.

The Hamming distance is a distance metric for two arrays consisting of binary elements, and it just fits into our problem domain because decisions are Bernoulli-distributed random variables.
The range of the Hamming distance is the set $\{0, 1\}$ because the maximum value of the score survives as $1$ and others turn into $0$ in classification. 
Therefore the Hamming distance, in this case, means the top-1 accuracy, which completely equals to the logical $and$ operation of two distributions. 
The accuracy is a discrete distance metric; therefore, we need to transform the accuracy into continuous space in order to analyze as a loss function. 
We define the batch accuracy loss function $L_H$ that utilizes the Hamming distance consisting of two distinct planes:
\begin{align}
    \label{eq:hamming_continuous}
     L_H(\symgt_1,\symgt_2) = \frac{1}{N_\text{cls}}\sum_{i=1}^{N_\text{cls}} \operatorname{max}(0, \symgt_{1, i} + \symgt_{2, i} - 1),
\end{align}
where $\symgt_{j,i}\in\{0, 1\}$ denotes the $i$-th element of vector $\symgt_{j\in\{1,2\}}$. 

Convexity and Lipschitz-continuity are indicators of the feasibility of a loss function in convex optimization. 
Both two properties of the accuracy loss function can be shown.
\begin{theorem}
    \label{theorem:hamming_convex}
    $L_H$ is a convex function.
\end{theorem}

\begin{proof} 
    \label{proof:hamming_convex}
    The Hessian matrix of $L_H$ is a $2 \times 2$ zero matrix because it consists of planes.
    
    $\therefore L_H$ is a convex function because the Hessian matrix of it is positive semi-definite~\ref{lemma:convexity}.
\end{proof}

\begin{lemma} 
    \label{lemma:convexity}
    Given a function $f: \mathbb{R}^{N_{in}} \mapsto \mathbb{R}^{N_{out}}$ if $f$ is semi-positive definite $\Rightarrow f$ is a convex function.
\end{lemma}

\begin{theorem}
    \label{theorem:hamming_lipschitz_continuous}
$L_H$ is a 1-Lipschitz-continuous function.
\end{theorem}

\begin{proof}
    \label{proof:hamming_lipschitz_continuous}
    \begin{equation}
        \label{eq:proof_hamming_lipschitz_continuous}
        \nabla L_H = 
        \begin{cases}
            \mathbf{0}, & \text{if } \symgt_{1, i} + \symgt_{2, i} < 1, \\
            (1, 1), & \text{otherwise,}
        \end{cases} \\
    \end{equation}
    where $\nabla$ denotes the Jacobian of a matrix. The slope of a straight line passes through two points on $L_H$ lies between 0 and 1 (inclusive).
    
    $\therefore L_H$ is a 1-Lipschitz-continuous function.
\end{proof}

Therefore, it is plausible that the accuracy loss function can be optimized using SPSA-based optimization.

\subsection{SPSA with Adaptive Momentum}
\label{ssec:methods-spsa_am}

SPSA is a robust algorithm, but this controls the magnitude of updates only based on the loss value without any corrections. 
We formulated SPSA with Adaptive Momentum~(SPSA-AM), which is a combination of SPSA and Adam optimization algorithms.
Decision-based attack using SPSA-AM is illustrated in Algorithm~\ref{alg:uap_optimization}.
We initialize the first perturbation $\sympert=\mathbf{0}$ and then randomly sample $\symstep \in \{\pm \delta\}^{N_\text{in}}$.
Then, $\symstep$ serves as the step in the optimization process.
We have two different perturbations: $\sympert^+$ and $\sympert^-$, by adding and subtracting the same step $\symstep$ to $\sympert$.
However, this does not guarantee $\|\sympert^+\|_l, \|\sympert^-\|_l \le \epsilon$. 
Therefore, we clipped the perturbation using the following equation:
\\
\begin{equation}
    \label{eq_clipping}
    \operatorname{clip}(\sympert) = \epsilon \frac{\sympert}{\|\sympert\|_l}, \quad l \in \{ 2, \infty \}.
\end{equation}
\\
And build the adversarial examples $\syminput^+$ and $\syminput^-$ with two opposite directions $\sympert^+$ and $\sympert^-$ are added to $\syminput$.
We calculate the pseudo-gradient $\symgrad$ using the decision of the two perturbations and a batch of images. 
The gradient $\symgrad$ does not require the decision of the clean images. 
Note that we clamped the adversarial examples using the lower and upper bounds of the original images. 
Finally, we update $\sympert$ using $\symgrad$ and step size $\alpha$.
We clipped the perturbation again because it does not guarantee the budget constraint.
The algorithm returns the final UAP $\sympert_\text{uni}$ after updating for all batches. 

\subsection{Extensions to Other Attack Tasks}
\label{ssec:methods-spsa_am}

\textbf{Targeted Attack.} We have addressed non-targeted attacks, so we did not specify the target category. 
While non-targeted attack aims to fool the victim, targeted attack aims to make the victim misclassify as a specific category. 
In other words, the accuracy loss function should be modified to count decisions not equal to the target. 
Therefore, the target accuracy loss function $L^\text{target}_H$ can be formulated as: 
\begin{equation}
    \label{eq:target-acc}
    L^\text{target}_H(\symgt_1,\symgt_2) = \frac{1}{N_\text{cls}} \sum_{i=1}^{N_\text{cls}}(1-\operatorname{max}(0, \symgt_{1, i} + \symgt_{2, i} - 1)).
\end{equation}

\noindent
\textbf{Score-based Attack.} Decision-BADGE is a way to build a decision-based universal adversarial perturbation using an accuracy loss function.
However, score-based universal adversarial perturbations can be generated using the same scheme by simply not applying $\mathbb{D}$ to $\symclassifier(\syminput)$.
Score-based attacks are discussed in more detail in the Comparison of Loss Functions section.

\begin{algorithm}[t]
    \caption{Decision-BADGE Algorithm} 
    \label{alg:uap_optimization}
    \textbf{Input:} $X,Y \gets \symniters$ input, ground truth batches \\
    \textbf{Parameters:} $\beta_1=0.5, \beta_2=0.999, \eta=10^{-8}, \\ \delta=0.01, \symgamma=0.001$\\
    \textbf{Output:} $\sympert_\text{uni} \gets$ optimized UAP
    \begin{algorithmic}[1]
        \State $\sympert \gets \mathbf{0}$
        \State $\mathbf{m}_0 \gets \mathbf{0}, \mathbf{v}_0 \gets \mathbf{0}$ \Comment {initialize moment vectors}
        
        \For{$t$ \textbf{in} $1$ \textbf{to} $\symniters$}
            \State $\syminput, \symgt \gets X_t, Y_t$ \Comment{iterating over batches}  
            \\
            \State $\symstep \gets \operatorname{random-select}(-\delta, \delta)$
            \State $\sympert^-, \sympert^+ \gets \operatorname{clip}(\sympert - \symstep), \operatorname{clip}(\sympert + \symstep)$
            \State $\syminput^-, \syminput^+ \gets \syminput + \sympert^-, \syminput + \sympert^+$ 
            \State $\syminput^- \gets \operatorname{clamp}(\syminput^-, \operatorname{min}(\syminput), \operatorname{max}(\syminput))$ \Comment{clamp $x_*$}
            \State $\syminput^+ \gets \operatorname{clamp}(\syminput^+, \operatorname{min}(\syminput), \operatorname{max}(\syminput))$ 
            \\
            \State $\hat{\symgt}^-, \hat{\symgt}^+ \gets \mathbb{D}(\syminput^-), \mathbb{D}(\syminput^+)$
            \State $\symgrad_t \gets \frac{L_H(\hat{\symgt}^-, \symgt) - L_H(\hat{\symgt}^+, \symgt)}{\symgamma}$  \Comment{apply $L_H$}
            \\
            \State $\mathbf{m}_t \gets \beta_1 \times \mathbf{m}_{t-1} + (1-\beta_1) \times \symgrad_t$ \Comment{apply Adam}
            \State $\mathbf{v}_t \gets \beta_2 \times \mathbf{v}_{t-1} + (1-\beta_2) \times \symgrad_t^2$
            \State $\hat{\mathbf{m}}_t, \hat{\mathbf{v}}_t \gets \mathbf{m}_t / (1-\beta_1^t), \mathbf{v}_t / (1-\beta_2^t)$
            \\
            \State $\sympert \gets \operatorname{clip}(\sympert + \alpha \symstep^{-1} \hat{\textbf{m}_t} / (\sqrt{\hat{\mathbf{v}}_t} + \eta))$
        \EndFor
        \State $\sympert_\text{uni} \gets \sympert$
    \end{algorithmic}
\end{algorithm}

\section{Experiments}
\label{sec:experiments}

In this section, we conduct extensive experiments on various datasets and victim models to evaluate the performance of Decision-BADGE in terms of attack success rate, the norm of the perturbation, and the training time.
We note that there are few methods for crafting UAPs in a real-world environment, so we included other methods in our experiments, such as white-box UAP and score-based SimBA. 
We also demonstrate the transferability and performance of the proposed method when applied to a targeted attack.
Furthermore, we analyze the effectiveness when applying various batch sizes, loss functions, and optimization algorithms.

\subsection{Experiment Settings}
\label{ssec:experiments-settings}

\begin{table*}
    \renewcommand{\tabcolsep}{1.15mm}

    \begin{subtable}[h]{\textwidth}
        \centering \small
        \begin{tabular}{l|cccc|cccc|cccc}
            \toprule
                                    & \multicolumn{4}{c|}{RN18 (95.48 \%)}                          & \multicolumn{4}{c|}{VGG19 (93.86 \%)}                                 & \multicolumn{4}{c}{MBN\_v2 (92.61 \%)}                            \\
                                    & \# UDT($\downarrow$)  & Time($\downarrow$)   & $l_2(\downarrow)$     & ASR($\uparrow$)  & \# UDT($\downarrow$)  & Time($\downarrow$)      & $l_2(\downarrow)$     & ASR($\uparrow$)       & \# UDT($\downarrow$)  & Time($\downarrow$)      & $l_2(\downarrow)$     & ASR($\uparrow$)   \\
            \midrule
            DF-UAP & 4.42E+5               & 91,945  & \textbf{338.45}    & 73.99            & 6.31E+5               & 132,863    & 380.43    & 29.36                 & 6.14E+5               & 226,809    & \textbf{276.58}    & 71.54             \\
            SimBA & 1.91E+7               & 120,206 & 393.52    & 78.25            & 1.98E+7               & 115,208    & \textbf{248.84}    & 49.44                 & 1.43E+7               & 150,357    & 410.87    & 81.79             \\
            DUAttack & 1.56E+5               & 28,852 & 541.77    & 85.10            & 1.56E+5               & 22,516 & 554.24    & 59.60                 & 1.56E+5               & 30,129 & 535.86    & 82.49             \\
            \textbf{D-BADGE (ours)} & \textbf{1.56E+5}      & \textbf{6,541}   & 409.05    & \textbf{85.88}   & \textbf{1.56E+5}      & \textbf{5,978}      & 425.73    & \textbf{63.53}        & \textbf{1.56E+5}      & \textbf{7,034}      & 398.89    & \textbf{88.06}    \\
            \bottomrule
        \end{tabular}

        \normalsize{\caption{Comparison among convolution-based architectures.}}
        \label{stab:exp_comparison_with_others_cnn}
    \end{subtable}

    \vspace{2.5mm}
    \begin{subtable}[h]{0.7\textwidth}
        \centering \small
        \begin{tabular}{l|cccc|cccc}
            \toprule        
                                    & \multicolumn{4}{c|}{ViT-T (62.45 \%)}                           & \multicolumn{4}{c}{Swin-T (77.59 \%)}                        \\
                                    & \# UDT($\downarrow$)  & Time($\downarrow$)      & $l_2(\downarrow)$  & ASR($\uparrow$)  & \# UDT($\downarrow$) & Time($\downarrow$)     & $l_2(\downarrow)$  & ASR($\uparrow$) \\
            \midrule 
            
            DF-UAP              & 3.72E+5               & 100,708     & 368.03 & 48.31            & 5.45E+5              & 342,664    & 314.64 & 30.91            \\
            SimBA            & 2.33E+7               & 173,816     & \textbf{117.27} & 23.36            & 2.60E+7              & 492,098    & \textbf{90.10} & 17.86            \\
            DUAttack & 1.56E+5               & 33,012 & 553.75    & 36.19            & 1.56E+5               & 36,249 & 554.24    & 42.66 \\
            \textbf{D-BADGE (ours)} & \textbf{1.56E+5}      & \textbf{9,890}      & 405.04 & \textbf{57.67}   & \textbf{1.56E+5}     & \textbf{11,010}    & 427.92 & \textbf{65.70}   \\
            \bottomrule
        \end{tabular}
        
        \caption{Comparison among transformer-based architectures.}
        \label{stab:exp_comparison_with_others_transformer}
    \end{subtable}
    \begin{subtable}[h]{0.3\textwidth}
        \centering \small
        \renewcommand{\tabcolsep}{5mm}
        \begin{tabular}{l|c}
            \toprule
            Architecture    & ASR   \\
            \midrule
            RN18            & 10.53 \\
            VGG19           & 10.13 \\
            MBN\_v2         & 24.11 \\
            ViT-T             & 7.02  \\
            Swin-T            & 10.98 \\
            \bottomrule
        \end{tabular}
    
        \caption{ASR using random perturbations.}
        \label{stab:exp_noise_fr}
    \end{subtable}

    \caption{Performance comparison with other methods. (a) ResNet18 (RN18), VGG19, MobileNet\_v2 (MBN\_v2), (b) Vision~Transformer (ViT-T) and Swin~Transformer (Swin-T) were tested as victim models. 
    UAP, SimBA, and DUAttack are the baselines. 
    The number inside the parenthesis refers to the accuracy of the victim model with clean inputs. 
    Note that DF-UAP is a white-box attack method and SimBA is an image-dependent attack method.
    (c) shows the ASR of perturbations with five different random noises. 
    The $l_2$-norms of the random noises were made equal to the $l_2$-norms of optimized perturbations.
    \textbf{Bold} font numbers indicate the best result in each evaluation metric.}
    \label{tab:exp_comparison_with_others}
\end{table*}

\textbf{Victim Models and Datasets.}
We evaluated ResNet18 (RN18), ResNet20 (RN20) \cite{He_2016_2016}, VGG19 \cite{Simonyan_2015_ICLR}, MobileNet\_v2 (MVN\_v2) \cite{Sandler_2018_CVPR}, ResNeXt29\_2x64d~(RNX29)~\cite{Xie_2017_CVPR}, Vision Transformer~(ViT-T)~\cite{Dosovitskiy_2021_ICLR} and Swin Transformer~(Swin-T)~\cite{Liu_2021_ICCV} for the CIFAR-10~\cite{Krizhevsky_2009_Toronto} dataset and a simple toy convolutional network for the MNIST~\cite{LeCun_1998_IEEE} dataset.
In the CIFAR-10 dataset experiment, we compared Decision-BADGE with DF-UAP, SimBA, and DUAttack attack methods. 
UAP, DUAttack, and Decision-BADGE used the CIFAR-10 training set to create universal perturbation, and SimBA built image-dependent perturbations using the validation set. 
All four methods were evaluated using the CIFAR-10 validation set. 
Additionally, we attacked the CIFAR-100 dataset, which is more challenging.
Unless otherwise stated, our experiments were primarily conducted using ResNet18.

\noindent
\textbf{Training Details.}
We controlled the step size using cosine annealing~\cite{Loshchilov_2017_ICLR} scheduling.
We had $\alpha$ to decay from $10^{-4}$ to $10^{-3}$ once over entire epochs. 
We decayed $\delta$ using step scheduling~\cite{Krizhevsky_2017_ACM} from 0.01 with a 0.9 decay ratio. 
$\gamma$ hyperparameter was set to $10^{-3}$. 
The default batch size was set to 256 by heuristic compromise. 
All training and evaluation samples are 8-bit images with values under 255.
The $l_\infty$-norm of perturbation was limited to 10.0.
All experiments were conducted on Ubuntu Server 18.04 with an Intel Xeon Gold 6226R 2.90GHz and NVIDIA RTX 3090.

\subsection{Evaluation Metrics}
\label{ssec:experiments-metrics}


\noindent
\textbf{Attack Success Rate (ASR, \%).}
ASR is our primary evaluation metric for non-targeted attacks, defined as the ratio of the number of changed decisions over the total number of adversarial examples. 
Similarly, we define target accuracy as an evaluation metric for a targeted attack. 
Target accuracy is defined as the number of adversarial examples classified to the targeted class.

\noindent
\textbf{Norm of Perturbation ($l_*$).}
The norm value of a perturbation is an intuitive evaluation metric for how the perturbation is recognizable by humans.
$l_{\infty}$-norm is the highest entry in the vector space. $l_{\infty}$-norm essentially determines the maximum magnitude of a given vector.
$l_{2}$-norm, the Euclidean or Frobenius norm, is the shortest distance between two vectors. 
It is calculated as the distance between the original and adversarial examples in the adversarial setting.
We primarily adopted $l_2$-norm to measure the overall distortion of adversarial examples.

\noindent
\textbf{The Number of Updates (\# UDT).}
It indicates how many times the perturbation requires to reach a certain ASR. 
The lower \# UDT suggests that the update directions to the perturbation were a better direction under a similar ASR.

\noindent
\textbf{Training Time (Time, seconds).}
Training time indicates how long a method takes to generate adversarial examples for the entire validation set. 
Therefore, it must be measured differently depending on image dependency. 
The perturbation optimization time could be the training time for optimizing universal perturbation and the time to build adversarial perturbations for all images in the validation set. 
The shorter training time suggests that the method can capture a victim's decision boundary in shorter periods of time.

\begin{figure*}[t!]
    \centering
    \begin{subfigure}{0.485\linewidth}
        \includegraphics[width=\textwidth]{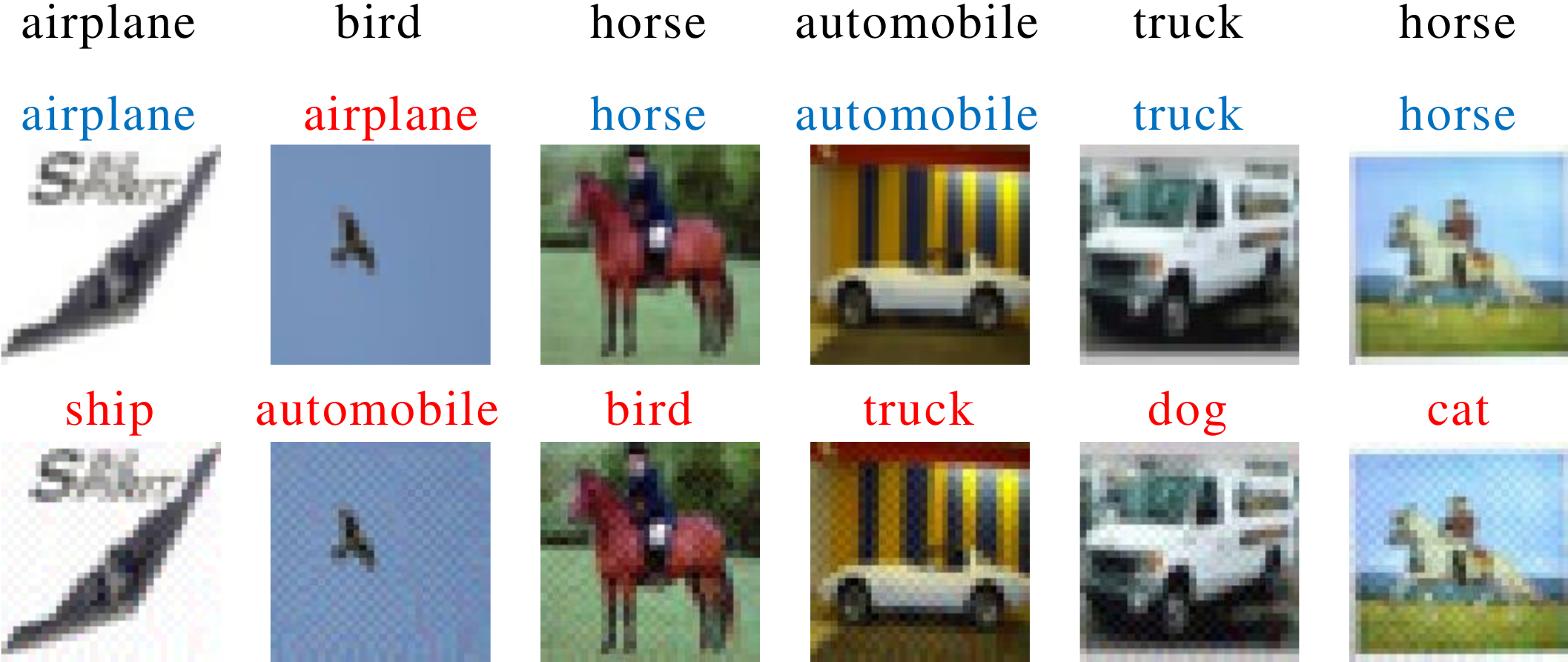}
        \caption{CIFAR-10}
        \label{subfig:examples_cifar10_nontarget}
    \end{subfigure} \hspace{0.7em}
    \begin{subfigure}{0.485\linewidth}
        \includegraphics[width=\textwidth]{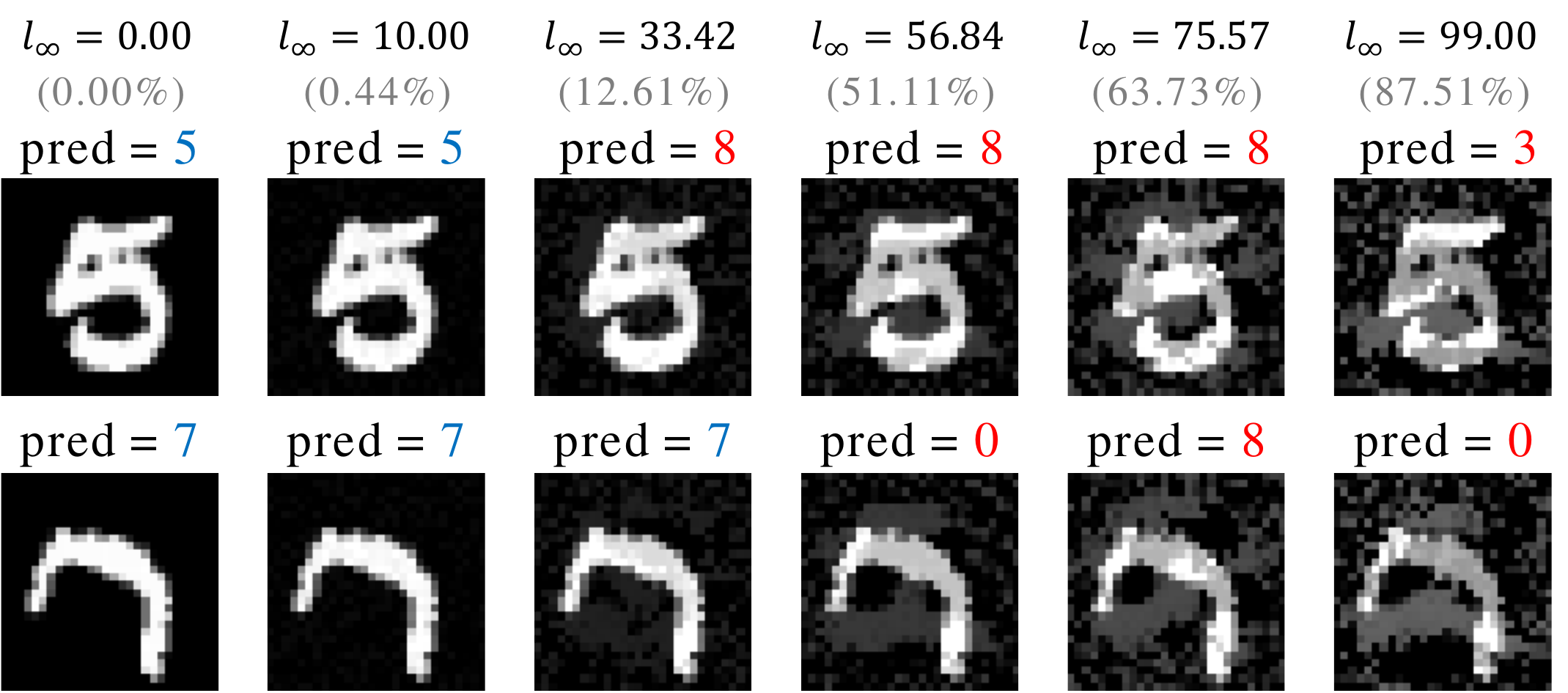}
        \caption{MNIST}
        \label{subfig:examples_mnist_budget}
    \end{subfigure}

    \caption{(a) shows the adversarial examples of the CIFAR-10 dataset against ResNet18. The top row illustrates the original images, accompanied by both the ground truth and the corresponding predictions. The bottom row shows the adversarial examples with their classification results. (b) shows the adversarial examples of the MNIST dataset on various $l_\infty$-norm constraints. The numbers in parentheses denote the ASR scores. Blue and red colors indicate correctly classified and misclassified categories, respectively, in both subfigures.}
    \label{fig:examples}
\end{figure*}

\subsection{Non-targeted Attack}
\label{ssec:experiments-non_targeted_attack}

We tested four attack methods: UAP, SimBA, DUAttack, and Decision-BADGE to five victim models: ResNet18, VGG19, MBN\_v2, ViT-T, and Swin-T on the CIFAR-10 dataset.
UAP and Decision-BADGE generated universal perturbations using a training set (50,000 images) and evaluated using the validation split.
All experiments were performed under $l_{\infty}=10.0$ constraint.
As shown in Table~\ref{tab:exp_comparison_with_others}, Decision-BADGE achieved higher ASR with fewer updates and shorter training time, even compared to the DUAttack with the same number of updates and queries.
The architecture of Vision Transformer and Swin Transformer significantly stand out from convolution-based networks.
Swin-T showed greater robustness against UAP, SimBA, and DUAttack compared to ViT-T, while Decision-BADGE outperformed others across the victims as shown in Table~\ref{stab:exp_comparison_with_others_transformer}.
We measured ASR on CIFAR-100 dataset against ResNet18 and Decision-BADGE works as well regardless of the number of categories.
For additional results, please refer to the supplementary material.
We conducted a Decision-BADGE attack on the MNIST dataset using a toy network. 
Results are shown in Figure~\ref{subfig:examples_cifar10_nontarget}. 
The high-contrast pixels require a higher $l_2$-norm limit, but the most substantial perturbation remains less visible than the real pixel value.
The $l_2$-norms of UAP and SimBA are less than that of Decision-BADGE.
This implies that Decision-BADGE has better learnability because weaker perturbation generally leads to poor ASR.

\subsection{Transferability}
\label{ssec:experiments-transferatiblity}

\begin{figure}[t!]
    \centering
    \includegraphics[width=1\linewidth]{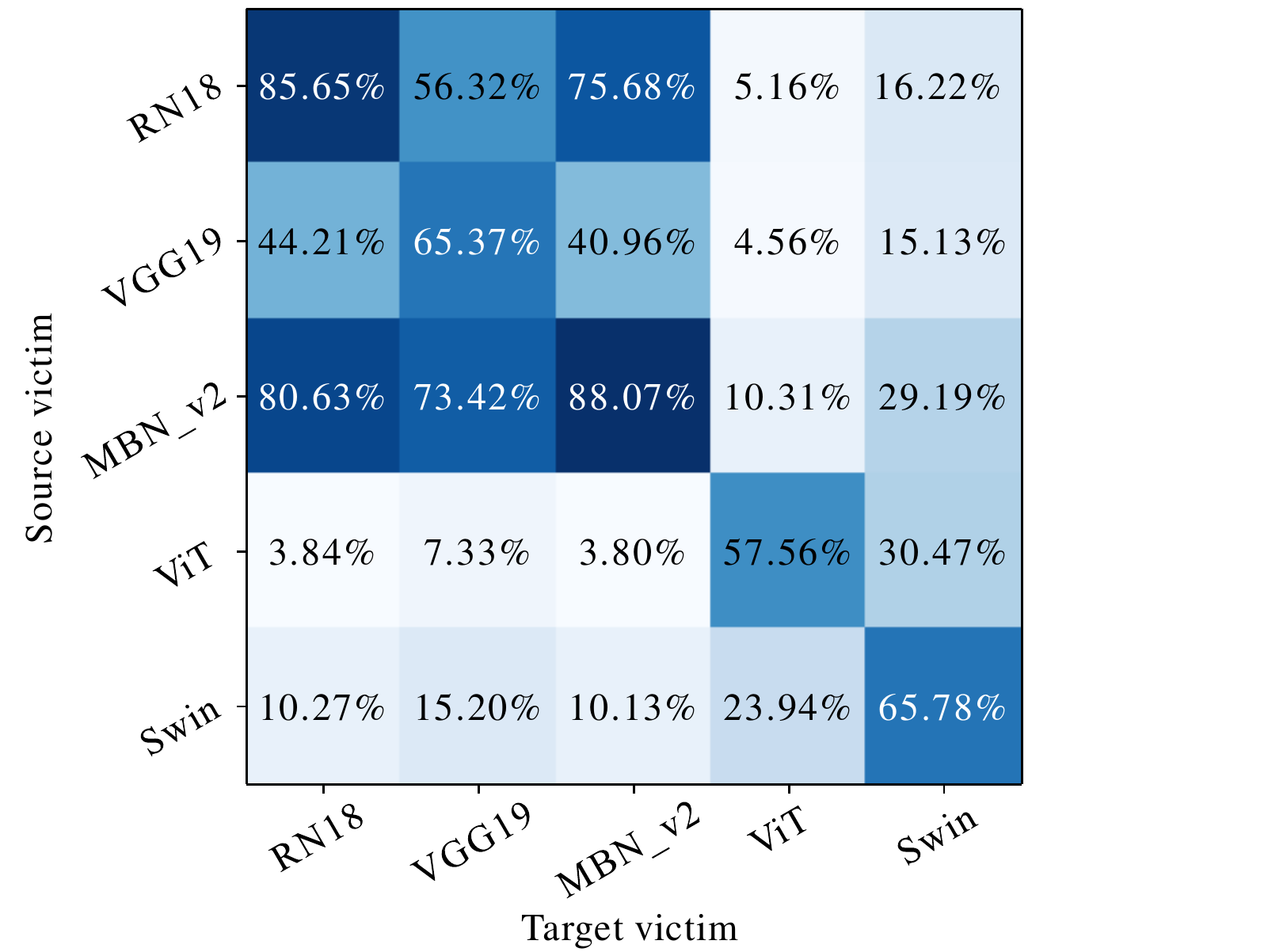}
    
    \caption{
    The confusion matrix of the transferability on non-targeted attack.
    The value of each cell refers to the ASR when fooling the target victim using a perturbation that was trained from the source victim. 
    }
    \label{fig:transferability}
\end{figure}

We investigated the transferability across victim models of the proposed methods.
In Figure~\ref{fig:transferability}, we demonstrated the transferability of our perturbations between various models. 
The figure summarizes the ASR of the Decision-BADGE attack trained on one network and evaluated on another.
Decision-BADGE is transferable within CNNs and transformers but not between a CNN and a transformer.
The ASRs are greater than 40\% among CNNs, which is much greater than the ASR of random noise with $l_\infty\text{-norm}=10.0$.
The ASRs among transformers are also greater than 20\%. 
However, the ASRs between a CNN and a transformer remain average of 10.923\%. 
CNNs and transformers are independent of each other, especially ViT-T.
Swin Transformer is relatively more transferable to CNNs and vise-versa with an average ASR of 16.02\%.
In summary, Decision-BADGE is effectively transferable among CNNs and transformers and partially among CNNs and transformers.

\subsection{Targeted Attack}
\label{ssec:experiments-targeted_attack}

We conducted a targeted Decision-BADGE attack on five victims that were trained using the CIFAR-10 dataset. 
The results are displayed in Table~\ref{tab:exp_target_accuracy}. 
Decision-BADGE achieved higher target accuracy for a specific class within each victim architecture. 
Specifically, MVN\_v2 outperformed the others, achieving the highest ASR target accuracy.
VGG19 exhibited poor ASR in non-targeted attacks, yet specific target classes were still attainable with VGG19.
The average target accuracy scores for all victims ranged from 70 to 85\%, indicating successful attacks.
The categories \textit{Plane, cat}, and \textit{frog} were easily targeted, while \textit{deer, dog}, and \textit{ship} proved to be robust in specific victim architectures. We discovered that the existence of classes with robust features can make adversarial attacks more challenging.

\begin{table}
    \centering \footnotesize
    \begin{tabular}{l|ccccc}
        \toprule
        Class         & RN18           & RN20           & VGG19          & MBN\_v2        & RNX29          \\
        \midrule
        Plane	& 85.94 & 88.72	& 93.93	& \textbf{97.45}	& 81.99 \\
        Car	    & 72.99 & 86.91	& 67.55	& \textbf{87.97}	& 86.36 \\
        Bird    & 87.66 & 56.21	& \textbf{95.03}	& 94.81	& 90.41 \\
        Cat	    & 84.57 & \textbf{95.00}	& 86.88	& 86.28	& 94.51 \\
        Deer	& 10.02 & 10.00	& \textbf{81.65}	& 77.72	& 77.44 \\
        Dog	    & 56.08 & \textbf{83.71}	& 9.73	& 67.41	& 74.75 \\
        Frog	& 91.63 & 86.35	& 95.60	& \textbf{97.75}	& 91.46 \\
        Horse	& 79.78 & \textbf{97.52}	& 88.63	& 91.65	& 92.41 \\
        Ship	& 80.73 & 10.03	& \textbf{98.34}	& 97.22	& 88.70 \\
        Truck	& 50.88 & \textbf{91.96}	& 73.02	& 90.69	& 82.55 \\
        \midrule
        \textbf{Mean}    & \textbf{70.03} & \textbf{70.64}	& \textbf{79.04}	& \textbf{88.90}	& \textbf{86.06} \\
        \bottomrule
    \end{tabular}

    \caption{
    Target accuracy~(\%) of the targeted attack for each category in the CIFAR-10 dataset. 
    }
    \label{tab:exp_target_accuracy}
\end{table}

\begin{table}
    \centering \small
    \begin{tabular}{r|cccc}
        \toprule
        Batch size      & \# Epochs & ASR             & Time          & \# UDT  \\
        \midrule
        1               & 4         & 21.03           & 2,931          & 200,000 \\
        32              & 128       & 83.39           & 3,069          & 199,936 \\
        64              & 256	    & 86.97           & 4,132          & 199,936 \\
        128	            & 512	    & 85.99           & 5,485          & 199,680 \\
        \textbf{256}    & 1,025     & \textbf{87.71}  & \textbf{8,679} & 199,875 \\
        512	            & 2,061     & 87.40           & 15,225         & 199,917 \\
        \bottomrule
    \end{tabular}

    \caption{Effectiveness of the batch size. The number of epochs~(\#~Epochs) was adjusted to make the number of updates as similar as possible.}
    \label{tab:exp_varying_batch_size}
\end{table}

\subsection{Effectiveness of Batch Attack}
\label{ssec:experiments-batch_attack}

We evaluated ASR across various batch sizes, adjusting the number of epochs to align with the number of updates to assess the batch size's effectiveness. 
We observed that as the batch size increases, the attack success rate also increases, given a similar number of updates, as shown in Table~\ref{tab:exp_varying_batch_size}.
However, this increased total training time, as it necessitated more inferences to the victim. 
In other words, a trade-off relationship exists between ASR and time in Decision-BADGE. Notably, the jump from 256 to 512 resulted in a dramatic increase in training time. 
We determined that 256 is the optimal batch size, considering a balance between ASR and training time compared to other sizes.

\subsection{Comparison of Loss Functions}
\label{ssec:experiments-loss}

We also conducted experiments on score-based attacks, assuming a score could be obtained by inserting an adversarial example into the victim model.
Four loss functions of Accuracy (ACC), KLD, CE, and EMD were tested. 
The result is shown in Table~\ref{tab:exp_varying_loss_function}.
ACC and KLD functioned better with decisions than scores, which is contrastive to CE and EMD. 
EMD worked poorly, especially with decisions.
It is because EMD, intuitively speaking, calculates the amount of data that must be transferred between two distributions to make them equal.
This is not a direct distance metric between two decision distributions. 
On the other hand, other functions are well-known distance metrics.
As a result, we obtained the best ASR with CE and ACC loss in score-based and decision-based attacks, respectively. 
The ASR using ACC on decision-based attacks is slightly lower than score-based attacks using CE but still performed with the lower variance of ASR.
This suggests that Decision-BADGE works, at least, as well as on score-based attacks despite of catastrophic lack of information.

\begin{table}
    \centering \small
    \setlength{\tabcolsep}{0.35em} 
    \begin{tabular}{l|cccc}
        \toprule
            & ACC                                   & KLD               & CE                                    & EMD               \\
        \midrule
        SB  & 72.96 $\pm$ 2.99                      & 87.69 $\pm$ 0.62  & \textbf{88.17} $\pm$ \textbf{0.45}    & 73.42 $\pm$ 1.72  \\
        DB  & \textbf{87.22} $\pm$ \textbf{0.29}    & 86.72 $\pm$ 0.32  & 86.73 $\pm$ 0.64                      & 2.67 $\pm$ 0.11   \\
        \bottomrule
    \end{tabular}

    \caption{Performance comparision for various loss functions. 
    Each value means $\mu \pm \sigma$ of ASR against ResNet18. 
    SB and DB refer to score-based and decision-based attacks respectively.}
    \label{tab:exp_varying_loss_function}
\end{table}

\subsection{Comparison of Optimization Algorithms}
\label{ssec:experiments-algs}

\begin{figure}[t!]
    \centering
    \includegraphics[width=0.9\linewidth]{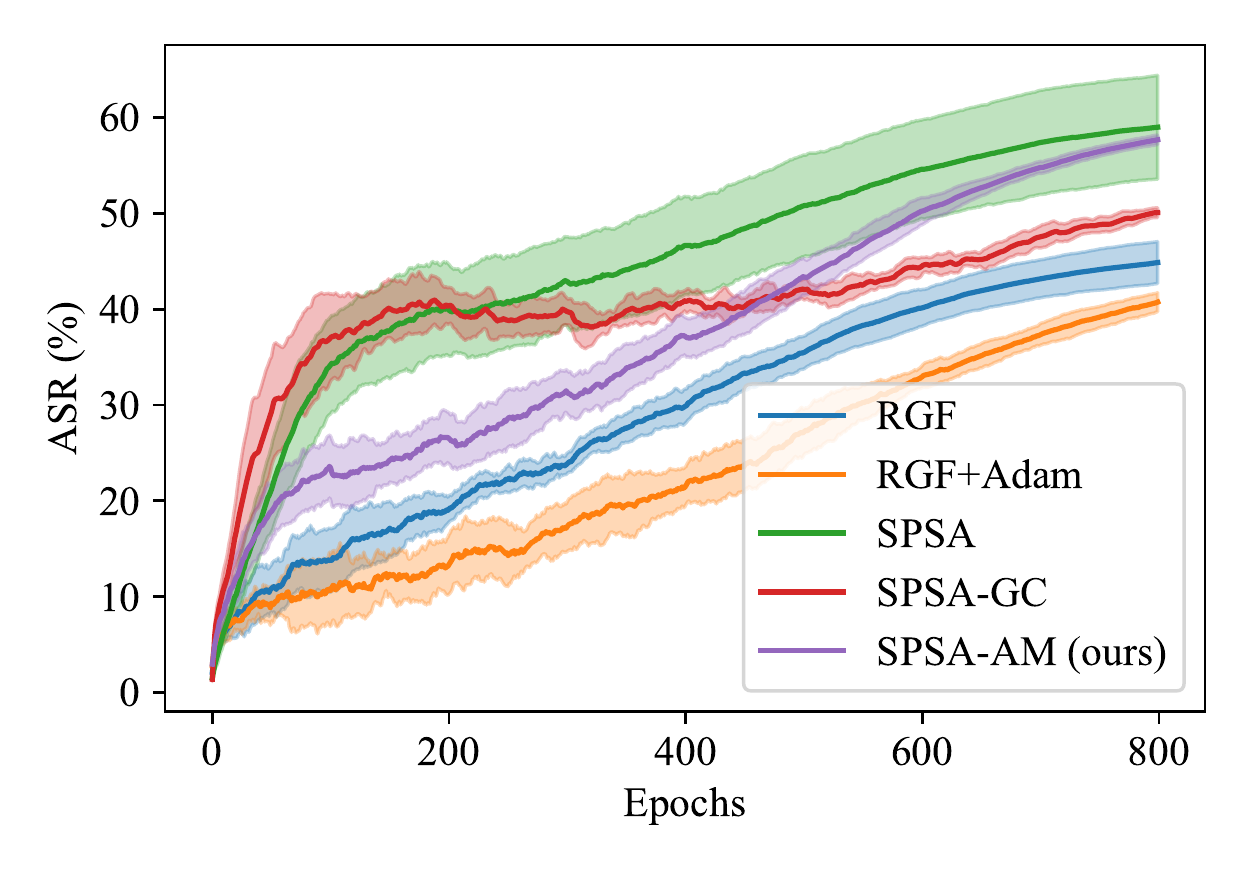}
    
    \caption{Convergence using different algorithms. Five different combinations of optimization algorithms were tested against ViT-T. 
    The ASRs of 20 attacks were averaged and the error range shows $2\sigma$ of them.}
    \label{fig:algorithms}
\end{figure}

The optimization algorithm is a critical factor in decision-based attacks.
As we combined the SPSA and the Adam optimizer, we also tested other combinations of optimization algorithms. 
We evaluated five combinations as shown in Figure~\ref{fig:algorithms}.
SPSA-based algorithms are basically performed better than RGF-based algorithms. 
Gradient correction using NAG or Adam helps stable convergence of perturbations.
We captured that SPSA-GC is more stable than SPSA while the ASR is much lower.
However, SPSA-AM still converges stably with a few ASR drops compared to SPSA.

\section{Conclusion}
\label{sec:conclusion}
This paper proposes a novel method named Decision-BADGE, which crafts image-agnostic universal perturbations in decision-based black-box attacks.
The proposed method utilize decisions of mini-batch to reconstruct probability distribution.
The accuracy loss function with SPSA-AM is used to optimize universal perturbation.
We demonstrated that the Decision-BADGE can easily be applied to targeted and score-based attacks.
The proposed method achieved better performance with less number of updates and training time compared to white-box UAP and score-based SimBA, even with the same number of updates and the same number of queries compared to DUAttack for CNN-based and Transformer-based victim models.
As it is evident that the number of queries is also a crucial factor in black-box attacks, we leave it as future work to craft UAPs with a small number of queries.

\section{Acknowledgement}
\label{sec:acknowledgement}
This work was supported by the National Research Foundation of Korea(NRF) grant funded by the Korea government(MSIT) (NRF-2022R1C1C1008074)

\bibliography{ms}

\end{document}


\maketitle

\section{$L_2$-norm Regulation}

\begin{figure}[h]
    \centering
    \includegraphics[width=1\linewidth]{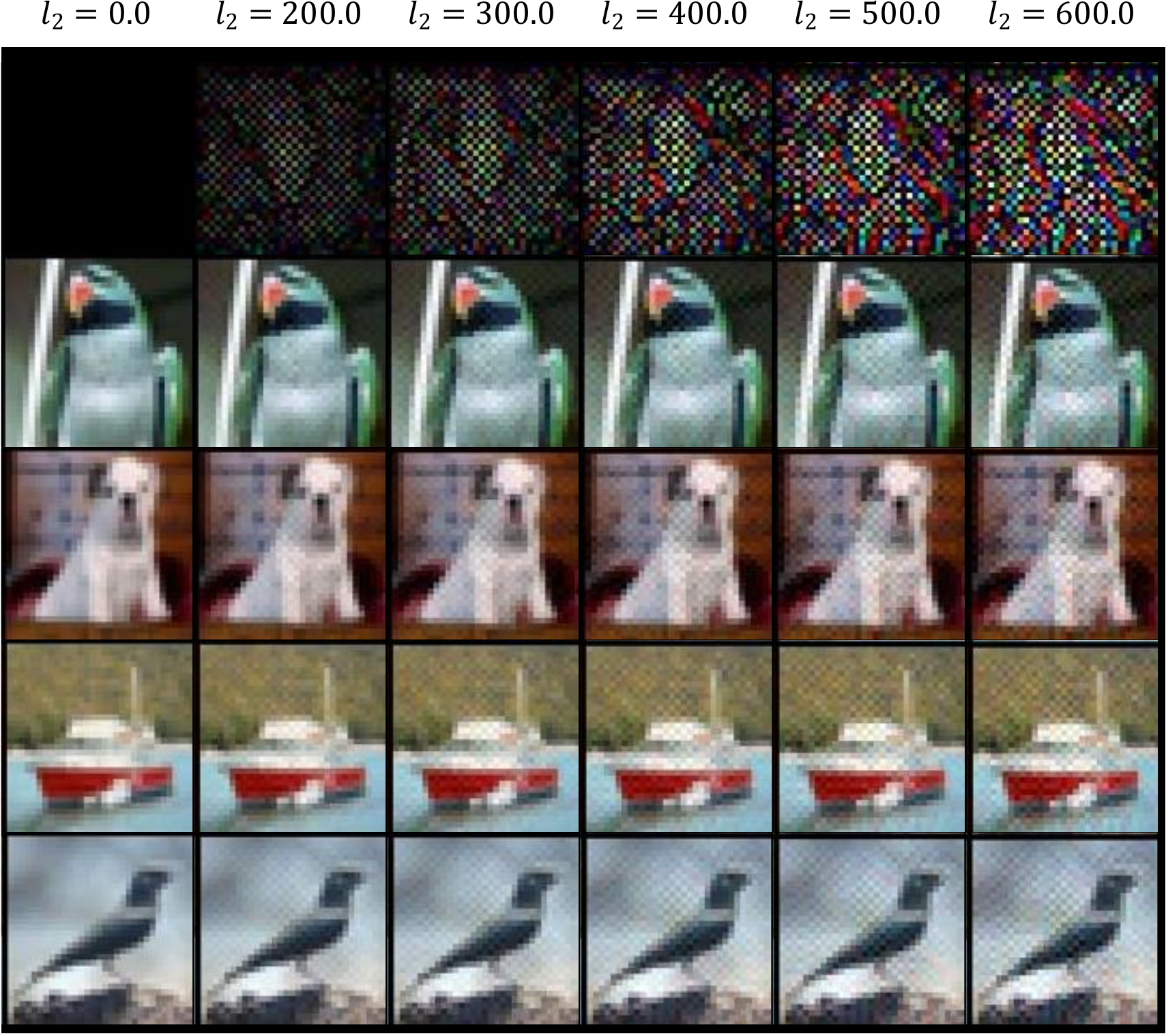}

    \caption{$L_2$-norm regulated perturbations and adversarial examples were visualized in this figure. The pixel values at the top row are scaled for visibility.}
    \label{fig:l2_vis}
\end{figure}

We regulated the maximum value of a pixel by $L_\infty$-norm by default. 
$L_\infty$-norm regulates only the maximum value, not the whole perturbation magnitude. 
$L_2$-norm can be another option of regulation in this context. 
We show how $l_2$-regulated perturbations look like and analyze their ASR.
As shown in Figure \ref{fig:l2_vis}, $l_2$-regulated perturbations have a check pattern similar to $l_\infty$-regulated perturbations.
Perturbations with higher $L_2$-norm achieved higher ASR as shown in Table \ref{tab:exp_l2}.
The growth is monotonic but not so fast at $l_2 \ge 400.0$ compared to at $l_2 \le 300.0$.

\begin{table}[h]
    \centering \small
    \begin{tabular}{c|cc}
        \toprule
        $l_2$   & $l_\infty$    & ASR (\%) \\
        \midrule
        200.0   & 11.48         & 53.63    \\
        300.0   & 16.66         & 74.81    \\
        400.0   & 24.58         & 83.66    \\
        500.0   & 27.95         & 87.46    \\
        600.0   & 36.17         & 88.47    \\
        \bottomrule
    \end{tabular}

    \caption{The magnitude of perturbation was regulated using $L_2$-norm and the ASR was measured.}
    \label{tab:exp_l2}
\end{table}

\begin{table}[h]
    \centering \footnotesize
    \begin{tabular}{c|ccccc}
        \toprule
        $l_\infty$  & RN18  & RN20  & VGG19 & MBN\_v2 & RNX29 \\
        \midrule
        7.0         & 82.27 & 72.82 & 38.63 & 84.95  & 86.58  \\
        10.0        & 86.96 & 85.51 & 68.94 & 88.49  & 89.03  \\
        13.0        & 88.43 & 87.81 & 84.77 & 89.60  & 90.87  \\
        \bottomrule
    \end{tabular}

    \caption{Evaluation of five victim architectures under different budgets on CIFAR-10 dataset.}
    \label{tab:exp_budget_success_rate}
\end{table}

\begin{table}[h]
    \centering \small
    \begin{tabular}{l|ccccc}
        \toprule
        \multicolumn{6}{c}{MNIST}                          \\
        \toprule
        $l_\infty$ & 10.00 & 33.42 & 56.84 & 75.57 & 99.00 \\
        \midrule
        ASR        & 0.44  & 12.61  & 51.11 & 63.73 & 87.51 \\
        \bottomrule
    \end{tabular}

    \caption{Effectiveness of the budget on MNIST. Our toy convolutional network on the MNIST dataset was tested. Five different $l_\infty$-norm budgets are tested in this experiment.}
    \label{tab:mnist}
\end{table}

\section{Effectiveness of Budget Limitation}
\label{sec:effectiveness_of_budget_limitation}

We measured ASR varying the maximum magnitude ($l_\infty$-norm) of a perturbation.
We followed the budget pool of GAP experimentation~\cite{Poursaeed_2018_CVPR}. 
ASR increased as the budget limit gets looser for all victim models we tested and there was no distinct difference as shown in Table~\ref{tab:exp_budget_success_rate} and Table~\ref{tab:mnist}. 
We found that the VGG19 architecture is sensitive to budget limitations but others are relatively not.
Table~\ref{tab:exp_budget_success_rate} shows that Decision-BADGE achieved better ASR under loose $l_\infty$ budget limits.
We also tested Decision-BADGE on the MNIST dataset as shown in Table~\ref{tab:mnist}. 
It requires a higher $l_\infty$-norm to fool the victim network due to its high-contrast characteristic. 
However, because of this characteristic, the strongest perturbation still has low visibility compared to the real pixel value.

\begin{figure*}[ht!]
    \begin{center}
        \begin{subfigure}{0.3\linewidth}
            \includegraphics[width=\linewidth]{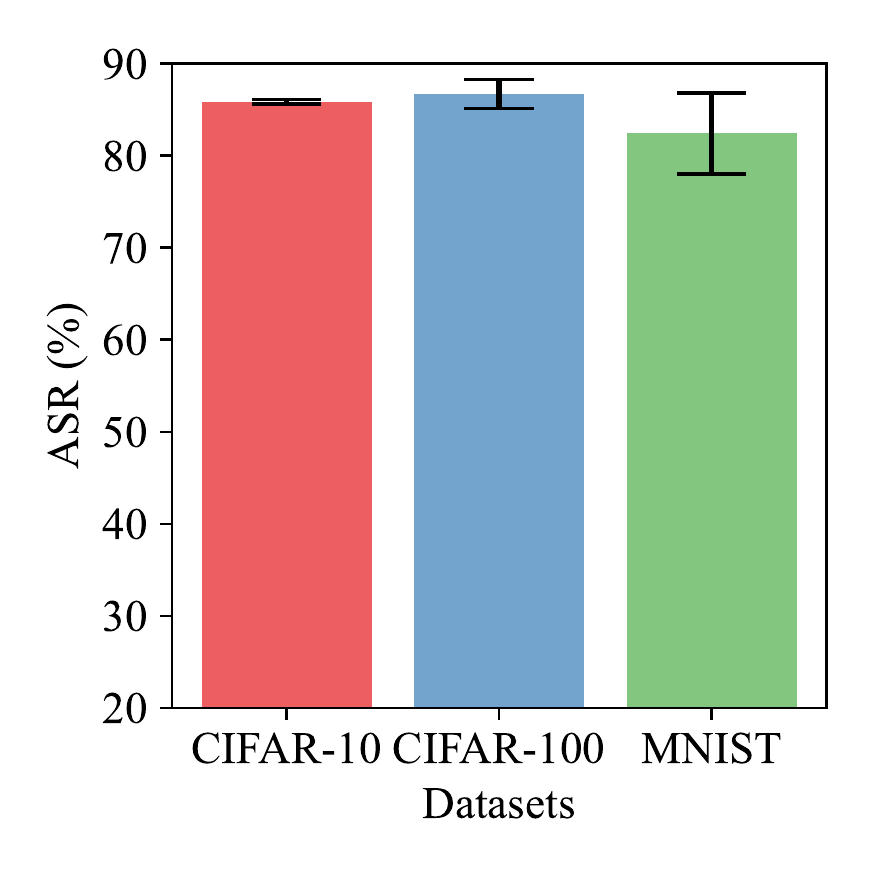}
            \vspace{-1.7\baselineskip}
            \caption{The mean and std of ASRs on datasets}
            \label{subfig:sig-dsets}
        \end{subfigure} \hspace{10mm}
        \begin{subfigure}{0.47\linewidth}
            \includegraphics[width=\linewidth]{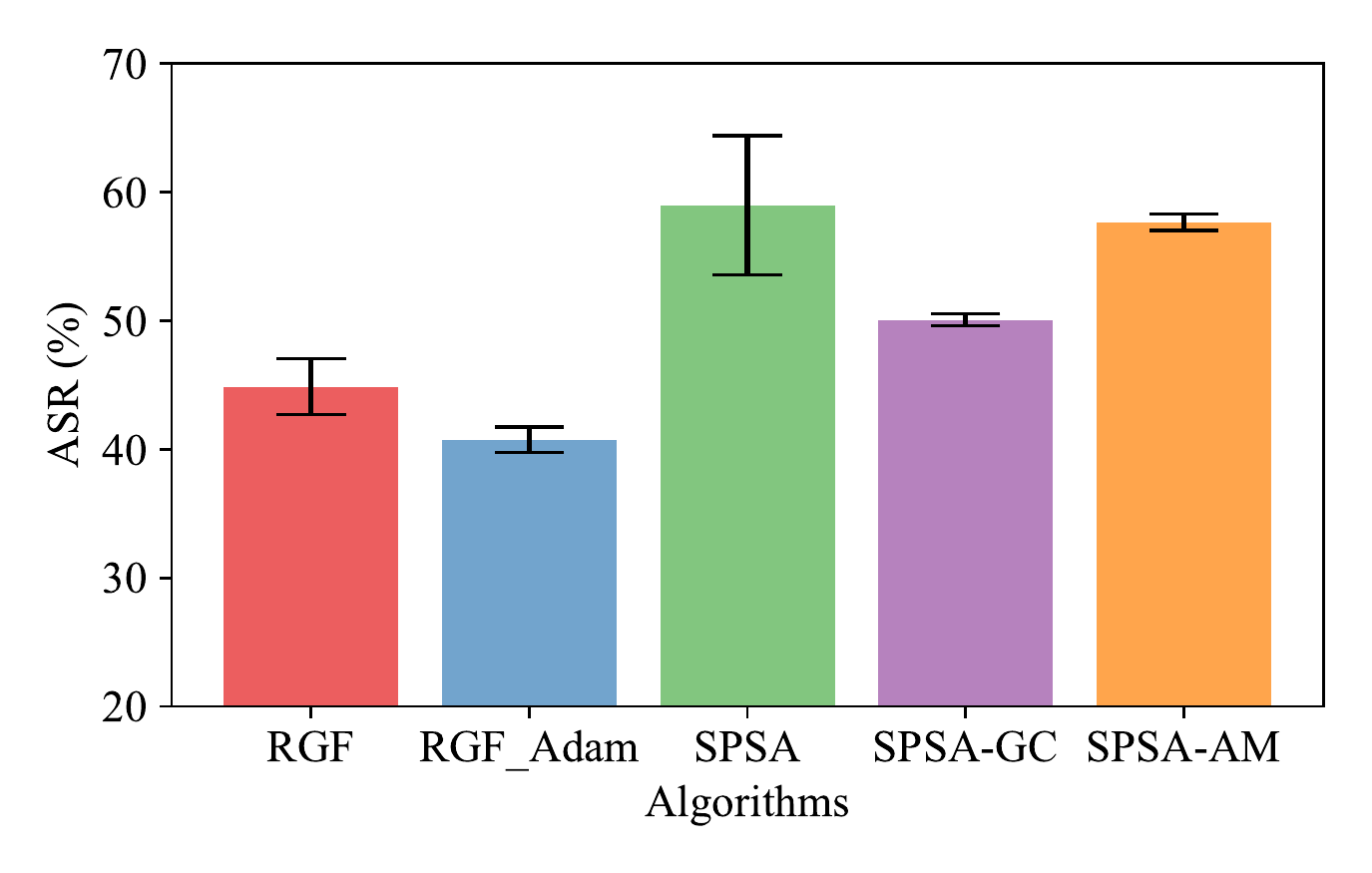}
            \vspace{-1.7\baselineskip}
            \caption{The mean and std of ASRs using different algorithms}
            \label{subfig:sig-algs}
        \end{subfigure}
    \label{fig:batch_attack}
    \vspace{-0.25\baselineskip}
    \caption{(a) shows the ASR on different datasets using ResNet18. The CIFAR-10 and CIFAR-100 datasets were not significantly different while MNIST is significantly different from CIFARs. (b) shows the ASR using different algorithms and ViT. SPSA-AM~(ours) shows the highest and most stable ASR. The error bars in both subfigures show the $2\sigma$ of the ASRs.}
    \end{center}
\end{figure*}

\section{Statistical Analysis}
\label{sec:statistical_analysis}

We performed Decision-BADGE on the CIFAR-100 which categorized the same images in the CIFAR-10 into 100 categories. 
Twenty perturbations were trained on both the CIFAR-10 and the CIFAR-100 datasets against ResNet18. 
Then we calculated the $p$-value using T-Test.
As a result, we obtained $p=0.022$ which is not significant with a threshold of 0.005.
On the other hand, the $p$-value from repeated measures-ANOVA~(RM-ANOVA) on three datasets: CIFAR-10, CIFAR-100, and MNIST was $1.71\times10^{-5}$.
This implies that Decision-BADGE successfully fooled the victim network in similar data domains regardless of the size of the dataset as shown in Table~\ref{subfig:sig-dsets}.

The optimization algorithm is our other contribution. 
We performed RM-ANOVA on five different optimization algorithms: the basic RGF~\cite{Nesterov_2017_Foundations-of-Computational-Mathematics}, RGF with Adam~\cite{Kingma_2015_ICLR}, the basic SPSA~\cite{Spall_TAC_1992}, SPSA-GC which is a combination of SPSA and NAG~\cite{Mopuri_2018_CVPR} proposed with BlackVIP~\cite{Oh_2023_CVPR}, and SPSA-AM which is the proposed method that combines SPSA and Adam. 
The difference among the five algorithms was significant with the $p$-value of 3.68E-18. 
The comparison is shown Table~\ref{subfig:sig-algs}.
All experiments for statistical tests were performed 20 times for reliability.


\section{Convergence Trajectory}
\begin{figure*}[ht!]
    \centering
    \begin{subfigure}{0.47\linewidth}
        \includegraphics[width=\linewidth]{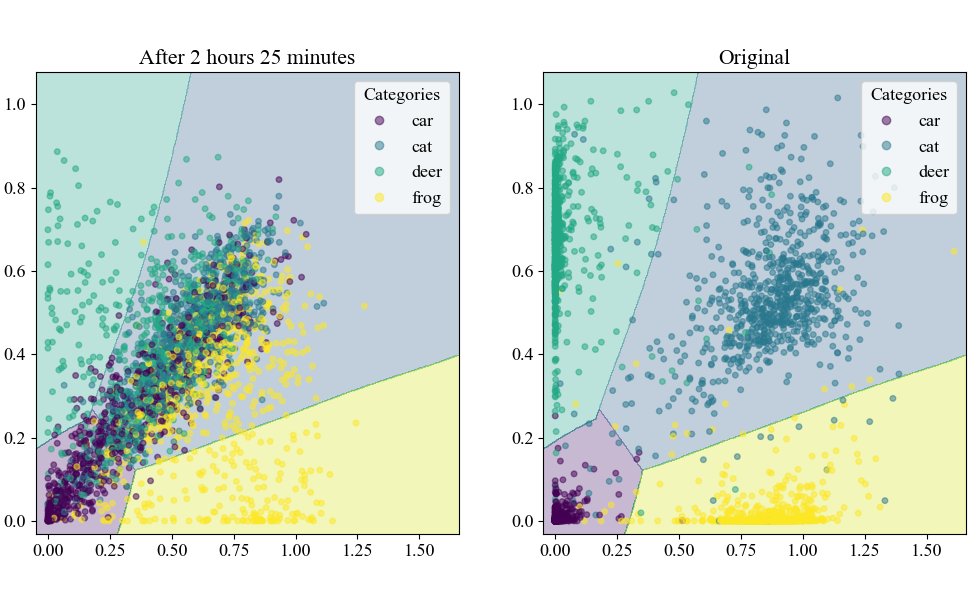} 
        \vspace{-1.7\baselineskip}
        \caption{Decision-BADGE}
        \label{subfig:badge_final_1000}
    \end{subfigure} \qquad
    \begin{subfigure}{0.47\linewidth}
        \includegraphics[width=\linewidth]{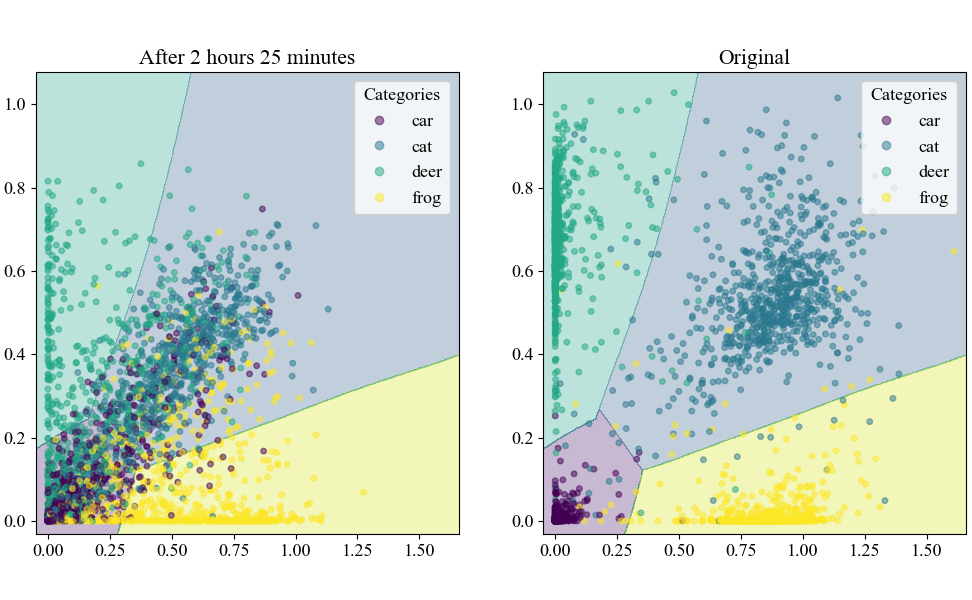} 
        \vspace{-1.7\baselineskip}
        \caption{UAP}
        \label{subfig:uap_final}
    \end{subfigure}
    \vspace{-0.25\baselineskip}

    \begin{subfigure}{0.47\linewidth}
        \includegraphics[width=\linewidth]{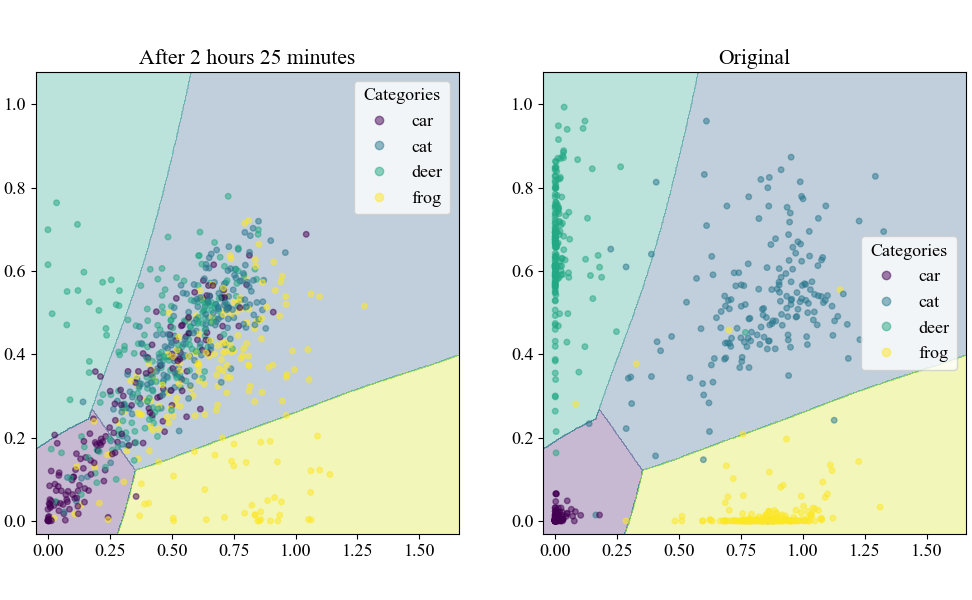} 
        \vspace{-1.7\baselineskip}
        \caption{Decision-BADGE (725 samples)}
        \label{subfig:badge_final_725}
    \end{subfigure} \qquad
    \begin{subfigure}{0.47\linewidth}
        \includegraphics[width=\linewidth]{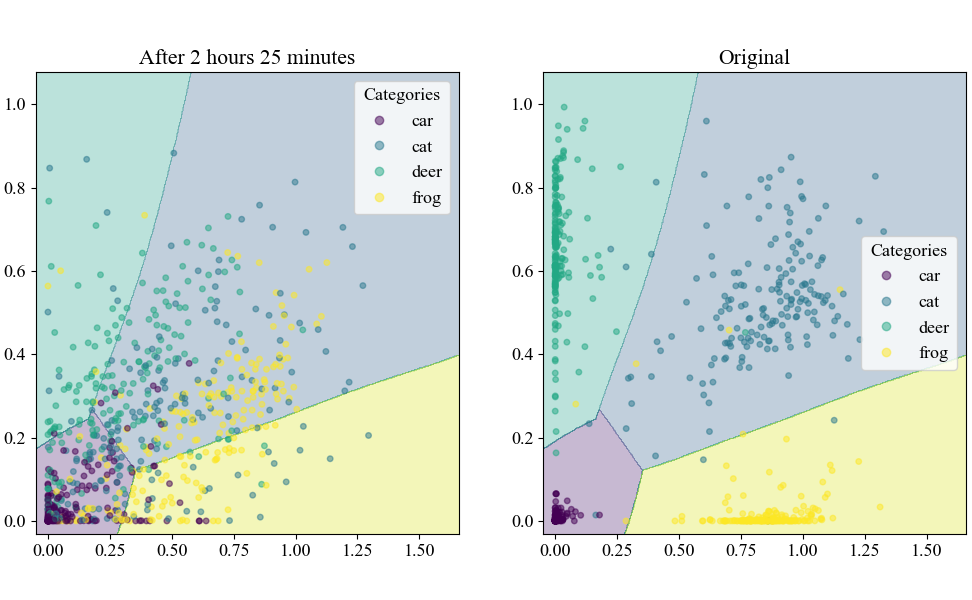} 
        \vspace{-1.7\baselineskip}
        \caption{SimBA (725 samples)}
        \label{subfig:simba_final}
    \end{subfigure}
    
    \vspace{-0.25\baselineskip}
    \caption{Each subfigure shows the datapoints on the decision space of a victim (ResNet18). The left-hand one is the adversarial datapoints and the other one is the clean datapoints without adversarial perturbations. We controlled the number of query points of (c) and (d) for clear understanding.}
    \label{fig:datapoints}
\end{figure*}

We compared the result of convergence efficiency (the fooling time, the number of updates) and attack success rate in the paper, which is briefly summarized in a table. 
On the other hand, visualizing how an attack method drives Gaussian noises to perturbations as static content is inappropriate.
Therefore, we provide a video that shows the trajectory, and Figure \ref{fig:datapoints} shows some screenshots of it.

Compared to UAP~\cite{Moosavi-Dezfooli_2017_CVPR}, which is a white-box image-agnostic attack method, optimizes a perturbation with one image at an update. 
Hence, each of the updates has a different direction.
On the other hand, each update of Decision-BADGE is performed with a batch of images, which makes the convergence direction stable.
Some adversarial datapoints of UAP stay at its original location while most of the datapoints of Decision-BADGE are perturbed.

SimBA~\cite{Chuan_2019_ICML} is another baseline which is an image-dependent black-box attack method. 
As shown in Figure\ref{subfig:simba_final}, SimBA successfully fooled most of datapoints.
But the perturbation of Decision-BADGE is capable of fooling other images while SimBA has to fool each of the others. 
That is time-consuming.

\bibliography{aaai24}